\documentclass[conference]{IEEEtran}
\IEEEoverridecommandlockouts
\usepackage{cite}
\usepackage{amsmath,amssymb,amsfonts}
\usepackage{algorithmic}
\usepackage{graphicx}
\usepackage{textcomp}
\usepackage{xcolor}
\def\BibTeX{{\rm B\kern-.05em{\sc i\kern-.025em b}\kern-.08em
    T\kern-.1667em\lower.7ex\hbox{E}\kern-.125emX}}
\begin{document}

\vspace{-2mm}
\title{LLM-Assisted Crisis Management: Building Advanced LLM Platforms for Effective Emergency Response and Public Collaboration}



\author{
\IEEEauthorblockN{Hakan T. Otal and M. Abdullah Canbaz}
\IEEEauthorblockA{\textit{
Department of Information Sciences and Technology
} \\
\textit{
College of Emergency Preparedness, Homeland Security, and Cybersecurity
}\\
\textit{
University at Albany, SUNY
}\\
Albany, NY, United States \\
hotal, mcanbaz [at] albany [dot] edu 
\vspace{-2mm}
}
}

\maketitle

\begin{abstract}
Emergencies and critical incidents often unfold rapidly, necessitating a swift and effective response. In this research, we introduce a novel approach to identify and classify emergency situations from social media posts and direct emergency messages using an open source Large Language Model, LLAMA2. The goal is to harness the power of natural language processing and machine learning to assist public safety telecommunicators and huge crowds during countrywide emergencies. Our research focuses on developing a language model that can understand users describe their situation in the 911 call, enabling LLAMA2 to analyze the content and offer relevant instructions to the telecommunicator, while also creating workflows to notify government agencies with the caller's information when necessary. Another benefit this language model provides is its ability to assist people during a significant emergency incident when the 911 system is overwhelmed, by assisting the users with simple instructions and informing authorities with their location and emergency information.

\end{abstract}


\vspace{-2mm}
\section{Introduction}

Urban centers like New York City pulsate with vibrant life, innovation, and economic might. Yet, beneath their dynamism lies a vulnerability woven into the fabric of their complexity: a susceptibility to large-scale disruptions. The United States alone grapples with an average of 50 major disasters annually, impacting millions and inflicting billions in damages \cite{urban_systems_vulnerability}. These emergencies, from natural disasters to infrastructure failures, hold the potential to devastate, as Hurricane Sandy tragically demonstrated in New York City, damaging over 69,000 residential units, leading to the temporary displacement of thousands of New Yorkers, and incurring immense economic losses\cite{nyc_hurricane_sandy_2023,pourebrahim2019understanding, wang2019vulnerable, hughes2014online}.

Traditional emergency response systems face hurdles in navigating the rapids of such crises. Information overload, communication bottlenecks, and the need for rapid, coordinated action within dynamic, often chaotic environments pose significant challenges \cite{loreti2022local,grace2021overcoming,misra2020information}. These difficulties are amplified in densely populated urban settings, where a single event can ripple through critical infrastructure and impact millions instantly \cite{andreassen2020information,el2019case}. Additionally, the linguistic diversity in urban areas, such as New York City, adds another layer of complexity. With a significant portion of 911 callers being non-native English speakers, the risk of miscommunication increases, further challenging the efficiency and effectiveness of emergency response systems\cite{clayman2023dispatching, dimou2021faster,mondal2021emergency}.




Integrating AI into the first responders' ecosystem holds immense promise, but navigating its intricacies demands careful consideration of sub-cuyltural and ethical currents \cite{adam2022mitigating, 10.1145/3375627.3375833, poursabzi2021manipulating}. Concerns from first responders, the FCC, and other stakeholders center around four critical pillars: generational techno-scepticism, potential biases, data privacy vulnerabilities, and the delicate balance of human oversight \cite{weiss2021business,national2007improving, buchanan2022role}.

The first obstacle to be overcome when planning the introduction of new technology to first responders is a culture that traditionally has been skeptical of technology \cite{pilemalm2022barriers}. Many older first responder leaders have grown up in a socio-technical culture in which "hoses and irons", street smarts and command intuition loomed larger than big data and advanced computational capabilities. Even those first responders who are digitally savvy or even digitally native may be wary of algorithmic biases creeping into AI-powered tools, potentially skewing data analysis or leading to discriminatory decision-making in critical situations \cite{andreassen2020information}. 

The FCC's main role during emergencies is to regulate telecommunications carriers, ensuring reliable and resilient communication services. This includes mandating carriers to offer text-to-911 services and promptly inform Public Safety Answering Points (PSAPs) of service impairments \cite{fcc_emergency_communications_2023}. The FCC also enforces outage reporting, strengthens e-911 connections, and convenes groups to develop best practices for emergency communications. Though not directly overseeing PSAPs, the FCC aids carriers in service restoration, like facilitating fuel supply for backup generators during crises. Additionally, the FCC regulates Wireless Emergency Alerts, focusing on message details and coverage accuracy \cite{fcc_wireless_emergency_alerts_2023}. Finally, stakeholders across the board acknowledge the crucial role of human expertise in guiding AI, advocating for clear oversight mechanisms to ensure responsible technology application and maintain human accountability in life-or-death scenarios\cite{sutton2021longitudinal, stern2022critical}. 

Although these challenges are significant, they are not necessarily prohibitive. Other technologies have been successfully introduced to first responders and good practices for so doing from this and other realms are known. Achieving ethical AI integration is possible with careful data management, ongoing efforts to reduce bias, and strong privacy protection methods \cite{harrison2018improving}. Additionally, creating clear and transparent guidelines for human-AI interaction is crucial for building trust and ensuring that both emergency responders and technology can function effectively in this sensitive environment \cite{sutton2021longitudinal, stern2022critical}. By giving priority to ethical issues and encouraging open communication, we can utilize AI not only to enhance efficiency but also to ensure responsible and equitable crisis management.

In the context of evolving technology, LLMs and generative AI mark a significant shift in crisis response capabilities. Their advanced language processing skills, developed through extensive parameterization and deep learning, go beyond basic content generation to address complex challenges in real-time emergency management \cite{rane2023contribution}. Consider the case of Hurricane Sandy, where the use of social media, particularly Twitter, was just the beginning for people seeking help and guidance. Now, LLMs have the potential to handle such situations more systematically. If designed and deployed with care, they can offer substantial benefits to all involved parties.

Recent events have further highlighted the need for advanced emergency response systems \cite{cdc_global_health_2023, dhs_fire_risk_2023, innovative_emergency_management_2023}. A poignant example is the earthquake in Turkey on February 6, 2023, which was a significant event with a profound impact on the country and its people \cite{chen2023early}. This earthquake, with a magnitude of 7.5, caused widespread destruction to buildings and infrastructure, leading to considerable loss of life and injuries. In the aftermath of the disaster, millions of people took to social media to tweet about the earthquake. Among them were individuals directly affected by the emergency and their relatives seeking help by explaining their situation with providing contact and location information. This surge in social media activity highlighted the urgent need for a more effective emergency response system capable of swiftly identifying and responding to such crises.

To this end, social media feeds and individual messages, previously a dense mass of information, can be converted into actionable insights by AI. LLMs can sift through large amounts of unstructured data to identify crucial information, such as distress signals hidden in digital chatter. This capacity to understand human language as it happens dramatically enhances situational awareness, aiding first responders in cutting through confusion with greater effectiveness. This is the real promise of AI in crisis management: transforming disparate calls for assistance into a clear picture of needs, directing response efforts with focused accuracy in challenging situations.

In this paper, we explore the transformative potential of foundation Large Language Models (LLMs) like LLAMA2 \cite{touvron2023llama} and Mixtral 8x7B \cite{mistralai_mixtral_8x7b} into emergency response systems. Our approach is to employ these AI models to interpret and act upon data from emergency calls and social media in real-time. Hence, we fine-tuned these open-source models to optimize the speed and effectiveness of crisis management, offering a dual benefit: enhancing the decision-support capabilities of first responders and providing real-time guidance to those affected by a huge crisis. 

As a means to this goal, leveraging the linguistic and analytical capacities of LLMs, we present the development of two frameworks: \textit{'Enhancing 911 Dispatch Efficiency with Large Language Model Integration'} and \textit{'LLMs for Collective Crisis Resolution: Public Collaboration'}. These frameworks, multi-modal and data-aware, focus on responsive and interactive emergency response. In the following sections, we elaborate on the technical specifications of the proposed systems, offer comparisons with various open-source models, address the challenges faced, and evaluate the potential of AI-driven solutions in managing emergencies within complex urban settings.

\vspace{-2mm}
\section{Enhancing 911 Dispatch Efficiency with LLM Integration}

The proposed framework integrates Large Language Models (LLMs) into critical communication systems, specifically tailoring this integration to enhance the 911 dispatch process. Central to this system is the real-time processing and analysis of emergency calls using LLMs, which operate alongside human dispatchers to improve response efficiency. This approach embodies a vision of synergistic collaboration between AI and human expertise, designed to bridge technological innovation and practical emergency management.

\begin{figure}[htbp]
\centerline{\includegraphics[width=\linewidth]{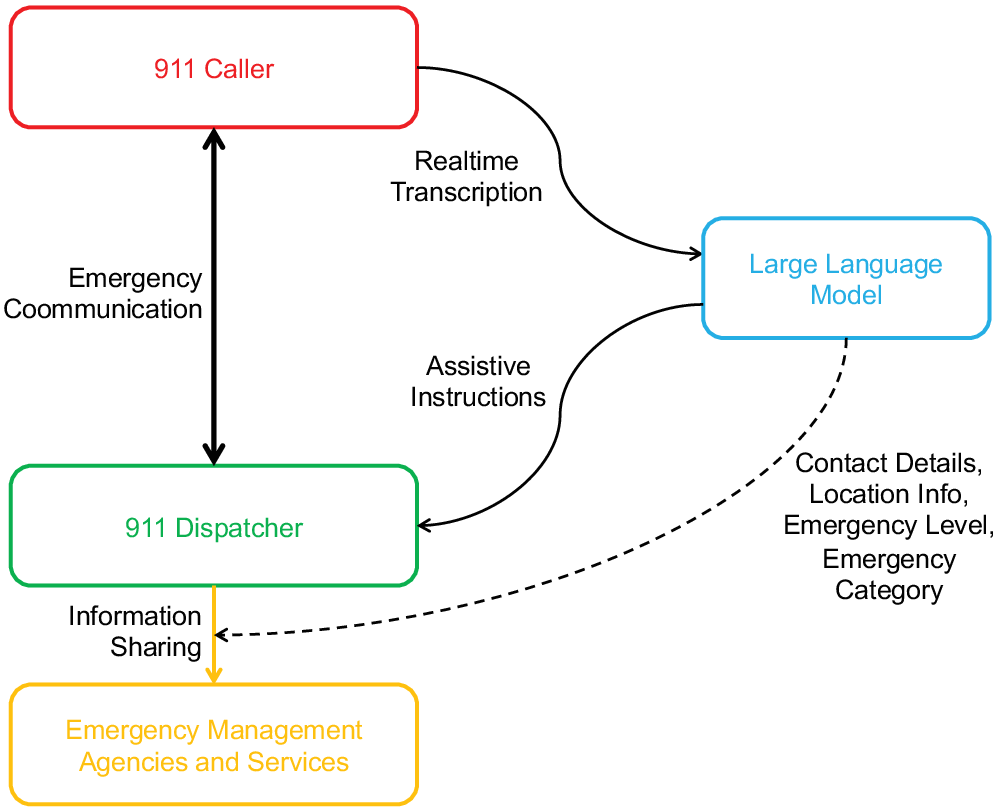}}
\vspace{-3mm}
\caption{Framework Design: Enhancing 911 Dispatch Efficiency with LLM Integration }
\label{fig:framework1}
\vspace{-2mm}
\end{figure}

Figure \ref{fig:framework1} illustrates the proposed system for integrating a Large Language Model (LLM) into the emergency response communication process, focusing on supporting the 911 dispatch process. The workflow begins when a 911 caller places an emergency call, which is received by a 911 dispatcher. Simultaneously, the call is also processed through an LLM, but it is crucial to note that the LLM does not interact directly with the caller.

Within the LLM pathway, an existing AI model performs real-time transcription of emergency calls, converting spoken language into text and conducting text segmentation for named entity recognition (NER). This NER process is crucial for identifying critical information such as emergency types, emergency levels, contact details, and other relevant data. This process includes multi-lingual support, a critical feature in diverse, populous cities such as New York City or Los Angeles. The transcribed text enables the LLM's analysis, leading to the generation of assistive instructions for the 911 responder or dispatcher. Additional details on LLMs used in this framework, including multi-lingual capabilities and performance comparisons, are provided in Section \ref{methodology}. Moreover, the LLM analyzes the transcribed text to extract vital information, including contact details, location, emergency level, and category. This information, marked by a dashed line in the documentation, indicates its suitability for forwarding to decision support systems. It assists dispatchers in filling out necessary forms, incident records, and facilitates communication with other parties involved in emergency management.

\vspace{-2mm}
\subsection*{Substantial Advantages}
The integration of LLMs into emergency response workflows enhances the accuracy and efficiency of incident management. The LLMs serve as a vital intermediary, improving communication between the caller and dispatcher and ensuring that critical information is promptly relayed to emergency management agencies. This demonstrates AI's potential in augmenting current emergency systems. Moreover, the progression of this technology involves integrating context and experiential learning. Unlike traditional computing systems, advanced models like ChatGPT, Bard, LLAMA, or Mixtral store information semantically, correlating language, concepts, and past events. This method resembles reinforcement learning, using historical experiences as a guide. For instance, dispatcher interactions with the LLMs, such as disregarding their recommendations, provide valuable feedback. This feedback teaches the LLMs about the relevance of their findings, prompting them to adjust future responses toward more meaningful and contextually appropriate directions.

Another dimension where LLMs can contribute significantly in the context of emergency response is in addressing the increasing challenges posed by language barriers. Between 1980 and 2019, the number of people in the U.S. who speak a language other than English at home nearly tripled \cite{us_census_bureau_language_use_2019}.  Consider a scenario where a dispatcher faces a language barrier with a caller. In such instances, LLMs can translate the caller's speech into the dispatcher's language. Furthermore, they can generate a set of straightforward questions in spoken language for the dispatcher to ask, thereby aiding in gathering more detailed information about the incident. LLMs, equipped with massively multilingual speech knowledge, can detect and operate in a wide range of languages, including those with limited data. This feature is particularly valuable in emergency situations where clear communication can be a matter of life and death. By integrating multilingual support, LLMs not only overcome language barriers but also ensure inclusivity and accessibility in crisis response.

Furthermore, LLMs can analyze transcribed calls and previous incidents to offer call-takers various response options for current situations, harnessing collective experience as the number of veteran call-takers may decline. Positioned as a virtual institutional memory, LLMs can be accessible to call-takers from day one. Additionally, as call-takers secure critical information about unfolding emergencies, LLMs can simultaneously develop a new workflow around this information, ensuring integration with other relevant law enforcement units, such as dispatch. This approach can further facilitate the more efficient and rapid mobilization of resources for emergency management.

\vspace{-2mm}
\section{LLMs for Collective Crisis Resolution: Public Collaboration}

Traditional emergency systems often face challenges in large-scale crises and emergencies, struggling to manage the high demand for assistance and information \cite{Welch22}. During major incidents like natural disasters or public health emergencies, the volume and complexity of requests can overwhelm these systems, causing response delays and miscommunications, sometimes leading to breakdowns in emergency management. Notable examples include Hurricane Sandy in 2012 and the Turkey Earthquake in 2023, which exposed the limitations of existing emergency systems. Consequently, people increasingly turn to social media to seek help and share information, making it a vital communication tool in distress situations.

The use of social media in emergencies, while transforming into a vital tool for reporting crises and seeking information, presents significant challenges.  The public nature of these posts raises serious privacy and security issues, as sensitive personal information can become exposed in widely accessible public domains. This situation underscores the need for controlled and expertly managed use of such platforms, potentially enhanced by the integration of Large Language Models (LLMs) for more effective emergency communication while safeguarding personal privacy.

\begin{figure}[htbp]
\centerline{\includegraphics[width=\linewidth]{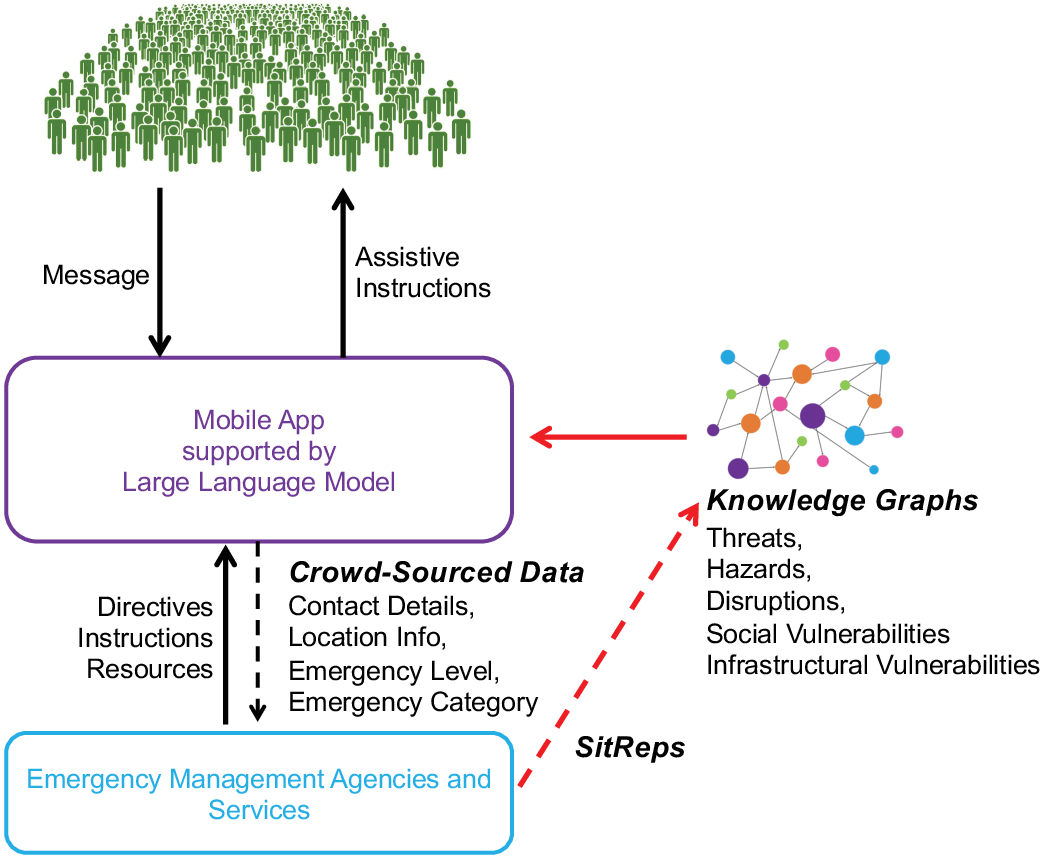}}
\vspace{-3mm}
\caption{LLM-Assisted Public Collaboration System Design}
\label{fig:crowd_flow}
\vspace{-3mm}
\end{figure}

To address these challenges, in our second framework, we propose an innovative system design, as illustrated in Figure \ref{fig:crowd_flow}, that leverages LLMs to assist the public during major emergencies. The system revolves around an LLM-enhanced mobile app, aiming to deliver AI-driven, real-time instructions and information to those impacted by crises. Managed by local emergency response centers, this app is tailored to establish a communication channel with first responders and offer immediate, pertinent advice to the crowds during emergencies. In pursuit of this, the LLM-enhanced mobile app, backed by the analytical intelligence of LLMs, can offer guidance on critical aspects such as safe shelter locations, provisions such as food and water, and where to wait for help. It can also provide personalized instructions based on the user's specific situation and location, which are automatically collected in users' messages via the application.

\vspace{-2mm}
\subsection*{Substantial Advantages}
Considering large-scale emergencies involving extensive populations, the primary functionality of the proposed framework is its capacity to systematically categorize various emergency types under authoritative oversight. For example, this system can classify incidents into groups, such as medical emergencies, fires, or infrastructure damage, and then relay this classified information to the corresponding local emergency response agencies. This systematic categorization enables a more structured and effective response from emergency response agencies.

As highlighted in the previously mentioned framework, the integration of multilingual support in LLMs not only helps overcome language barriers but also reinforces inclusivity and accessibility in crisis response.

Furthermore, the integration of LLMs into public emergency response systems offers a promising avenue for enhancing the efficiency and effectiveness of crisis management. By leveraging the power of AI to analyze the vast amount of data in seconds, categorize, and delegate the information of the emergency incidents to the corresponding authorities in a reverse crowd-sourcing manner, this framework can significantly improve the capacity of emergency services to handle large-scale crises. Moreover, by providing real-time, accurate information to the public, it empowers individuals to make informed decisions during emergencies, ultimately contributing to a more resilient and responsive emergency management ecosystem.

Please note that there exists a Text-to-911 service that enables text message communication with 911 emergency call takers, which is crucial for those with hearing or speech disabilities. It offers a critical alternative in scenarios where voice calls are impractical or unsafe. However, depending on the policies and decisions made by each individual emergency call center within the respective counties, the availability of this service may vary at the county level. Our proposed LLM-assisted framework is designed not to replace but to supplement and improve this existing communication infrastructure.

\vspace{-2mm}
\section{Methodology}
\label{methodology}

The methodology of this paper focuses on choosing and refining a suitable AI model with the help of a reliable, top-notch dataset. Crucial to training the LLM is a dataset that is diverse and accurately annotated, especially in the context of emergency response, where precision is paramount. The AI model must strike a balance between computational efficiency and nuanced language understanding to interpret various scenarios accurately. This synergy of a carefully curated dataset and an effective model is fundamental to achieving both accuracy and promptness in emergency response scenarios.

\vspace{-2mm}
\subsection{Model Selection}

In our investigation to identify the most suitable LLM for analyzing emergency situations, we focused on the LLAMA2 \cite{touvron2023llama} and Mistral \cite{mistralai_mixtral_8x7b} models. The LLAMA2 model, with its variants of 7B, 13B, and 70B, offers a scalable approach to language processing, capable of handling complex linguistic tasks with varying degrees of computational resources. The Mistral models, including the 7B and 8x7B versions, are also noteworthy open-source options known for being relatively smaller compared to LLAMA2 but proven to be effective in language understanding and processing tasks. These models were chosen for their proven performance and adaptability in complex language analysis.

\vspace{-2mm}
\subsection{Datasets}

Our research utilized two distinct datasets for fine-tuning the selected models. The first, the Turkey Earthquake X Corp. Dataset, comprises 500 tweets about the Turkey earthquake. These tweets are classified into two categories: Class 1, with 300 tweets from individuals directly affected by the earthquake seeking help and sharing location and contact details, and Class 0, including 200 tweets - 150 containing similar keywords to Class 1 such as "earthquake," "food," "blanket," etc., and 50 unrelated to the earthquake.

The second dataset, the Emergency-Disaster Messages Dataset, encompasses over 25,000 social media messages linked to various disasters, including the 2010 Haiti earthquake, the 2010 Chile earthquake, the 2010 Pakistan floods, and the 2012 Hurricane Sandy in the U.S.A. It covers a broad range of events and years, with non-English messages translated into English for uniformity. Please be aware that the data provider did the translation of these messages.

\begin{figure}[htbp]
\centerline{\includegraphics[width=1.05\linewidth]{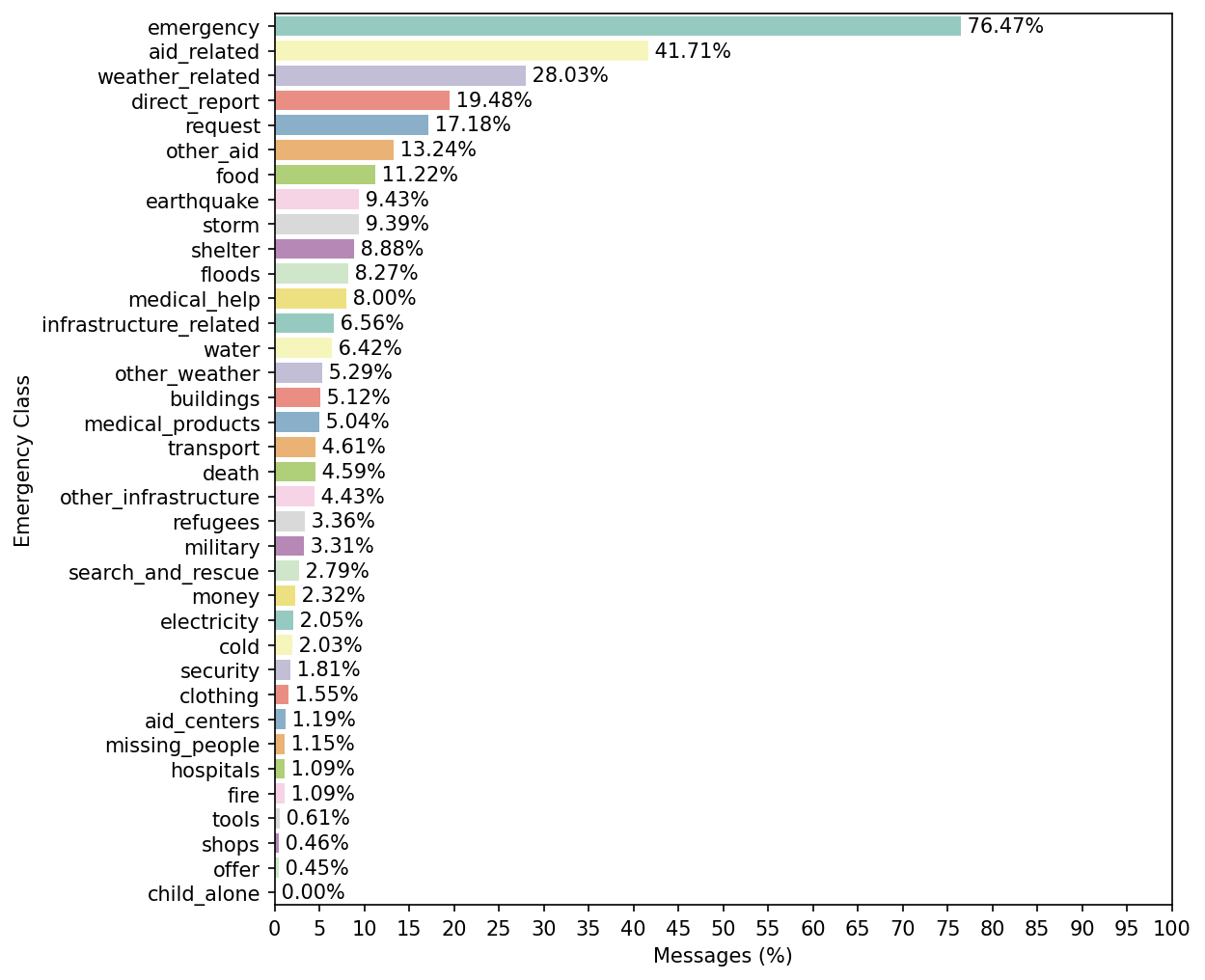}}
\vspace{-4mm}
\caption{Class Distribution in the Emergency-Disaster Messages Dataset}
\label{fig:data_distr}
\vspace{-3mm}
\end{figure}

In Figure \ref{fig:data_distr}, we chart the distribution of different types of emergency-related messages. The most common categories, 'emergency' and 'aid\_related,' account for 76.47\% and 41.71\% of messages, respectively, indicating a high frequency of requests for immediate aid. Following this are the categories 'weather\_related' and 'direct\_report,' which indicate that a sizable portion of messages are about weather-related issues and firsthand accounts of emergencies. Less frequent categories include requests for food, reports of earthquakes, and storm information. The least mentioned categories are related to offers of help, situations involving children alone, and inquiries about shops, indicating these are not as commonly discussed in emergency-related communications.

The data distribution highlights a significant imbalance, reflecting the complexity and challenge in curating emergency response datasets that require expert validation for message classification. This difficulty underscores the need for expanded and focused dataset development in future work. Our collection of earthquake-related data from Turkey's recent seismic event is a step towards addressing this gap, recognizing the crucial role of accurately labeled datasets by knowledgeable experts in improving emergency response systems.

\vspace{-2mm}
\subsection{Additional Resources}

In addition to the datasets, we incorporated training materials such as the FEMA handbooks, emergency response guidelines, and 911 dispatcher training manuals. These resources provide the LLM with context and guidelines for generating assistive advice, ensuring that the responses are not only accurate but also practical and in line with standard emergency response protocols.

\vspace{-2mm}
\subsection{Preprocessing \& Prompt Engineering}

The effectiveness of the LLMs in emergency response depends significantly on the preprocessing of the datasets and the engineering of prompts used for model training. The datasets, including the Turkey Earthquake X Corp. Dataset and the Emergency-Disaster Messages Dataset, underwent a thorough cleaning process. This involved removing special characters and emojis, filtering out irrelevant information, and discarding very short messages that lacked sufficient context. Such preprocessing ensures that the data fed into the LLM is relevant and clean, thus improving the model's learning efficiency and accuracy.

Prompt engineering is crucial in our approach, involving the craft of prompts that lead the model to effectively parse and classify emergency communications. We experimented with various prompt structures to refine both LLAMA2 and Mistral models' performances, using both multiclassification and binary classification techniques to determine the relevance and category of emergency messages. This process is instrumental in teaching the model to discern and accurately label emergency situations. The cleaned input texts, expected responses, and system prompts were combined to generate a comprehensive dataset suitable for fine-tuning LLAMA2.  By refining the input data and crafting precise prompts, we aimed to enhance the model's ability to provide accurate, relevant, and rapid assistance in crisis scenarios.

\vspace{-2mm}
\subsection{Supervised Fine-Tuning}

Supervised Fine-Tuning (SFT) is integral to training models like LLAMA2 and Mistral, particularly for models with extensive parameters. The pre-trained foundation models fine-tuned using the curated dataset to enhance their precision for emergency communication. We also utilized LoRA, a technique that allows only a small portion of the model to be trainable, thus reducing the number of learned parameters significantly. This allowed for efficient training by modifying a fraction of the models, thus reducing the number of parameters significantly and enabling the model to handle the intricacies of emergency scenarios.

The model's size and computational load are further optimized using Quantized Low-Rank Adapters (QLoRA), which sit atop a quantized, frozen model, preserving the base model's robustness. This fine-tuning process, which involves precise quantization, leads to a compact model without significantly affecting performance. The resulting model, which requires saving only the modifications, is compatible with various data types and retains the original model's integrity.

In enhancing the training process, techniques like NEFTune noise \cite{jain2023neftune} and Flash Attention 2 \cite{dao2023flashattention2} were incorporated to prevent overfitting and improve attention mechanism efficiency. By applying these methodologies to specific settings, the models are fine-tuned to efficiently and accurately process emergency communications, resulting in streamlined models adept at crisis response.

\vspace{-2mm}
\subsection{Hardware Requirements}

Training and deploying the LLAMA2 and Mistral models involved significant computational resources, leveraging the NVIDIA Base Command Cloud System for necessary processing capacity. We utilized NVIDIA A100 GPUs with 80GB VRAM for training the Llama2 13B model on over 20,000 texts. These GPUs facilitated quicker training and data management, essential for the real-time analysis required by our emergency response system. With methods like parallel processing, quantization and Flash Attention 2, training was expedited, culminating in a model that responds within two seconds, a critical factor for emergency scenarios. 

\vspace{-2mm}
\section{Experimental Results}

In our search for the ideal large language model for analyzing emergency situations, we assessed the LLAMA2 variants with 7B, 13B, and 70B parameters, alongside the Mistral models with 7B and 8x7B parameters. Our initial findings indicated that the 7B models, being smaller, provided quick processing times and handled simpler queries with high efficiency. Their compact size, however, did present limitations in complex understanding and contextual comprehension.

\begin{figure*}[!tb]
    \centering
    \begin{minipage}{.5\textwidth}
        \centering
        \includegraphics[width=.9\linewidth]{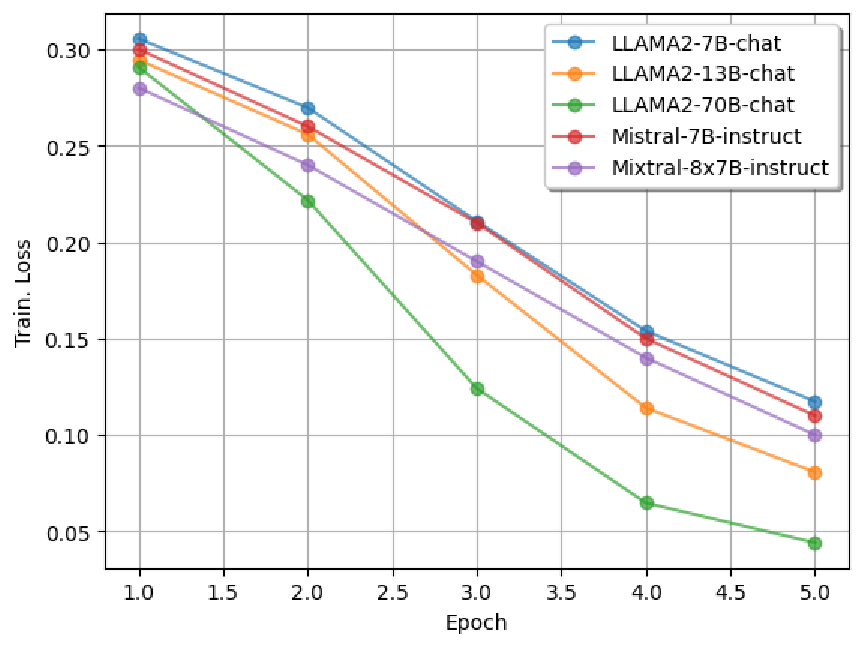}
        \vspace{-3mm}
        \caption{Training losses of different models over epochs}
        \label{fig:trainloss}
    \end{minipage}%
    \begin{minipage}{0.5\textwidth}
        \centering
        \includegraphics[width=.9\linewidth]{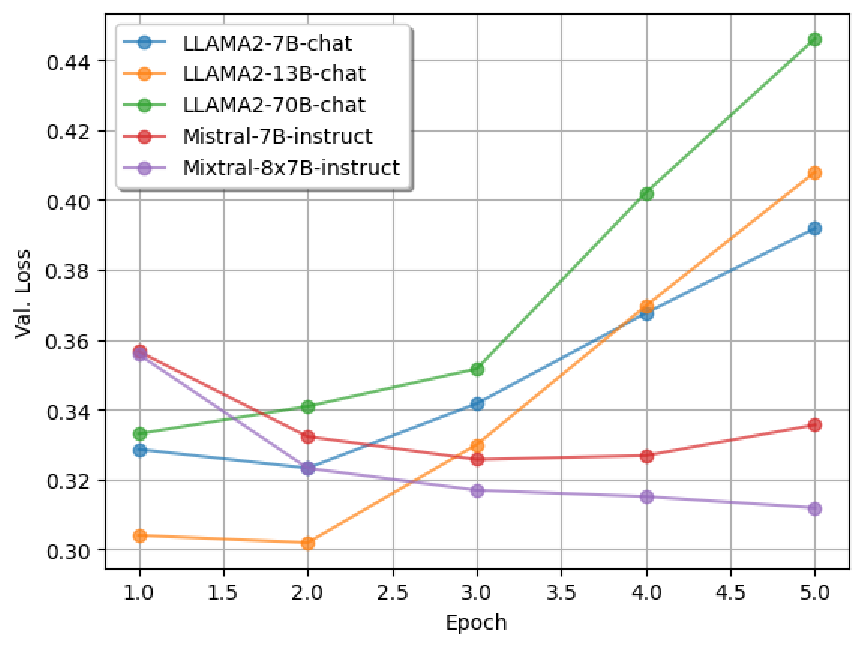}
        \vspace{-3mm}
        \caption{Validation losses of different models over epochs}
        \label{fig:valloss}
    \end{minipage}
    \vspace{-4mm}
\end{figure*}

The training losses, as illustrated in figure \ref{fig:trainloss}, reveal how each model learns to minimize errors over epochs. All models show a downward trend in training loss, indicating that learning is taking place effectively. Notably, LLAMA2-70B exhibits the steepest decline in training loss, suggesting a superior learning capacity likely due to its larger parameter count, which enables a deeper understanding of complex patterns in the data. However, it is also observed that the LLAMA2-70B model begins to plateau after epoch 4, which could indicate the beginning of overfitting or reaching its learning capacity.

In figure \ref{fig:valloss}, validation losses can be observed. Generally, LLAMA2 models initially perform better for 2 epochs, but as epochs increase, their losses rise, indicating potential overfitting to the training data. This highlights the challenge of finding a balance between model complexity and generalizability to new data. In contrast, Mistral 7B and 8x7B models show more stability in validation losses, suggesting better generalization despite their smaller size and indicating that they can be trained for more epochs.


\begin{table}[htbp]
\vspace{-2mm}
\renewcommand{\arraystretch}{1.2}
\caption{Metrics of the Tested Models}
\label{tab:metrics}
\centering
\begin{tabular}{|c||c|c|c|c|c|}
\hline
Model &  Prec. &  Recall &  F1 &  ROC-AUC &  Acc. \\
\hline\hline
LLAMA2-7B-chat        &       0.79 &    0.75 &      0.77 &     0.87 &      0.48 \\
LLAMA2-13B-chat       &       0.78 &    0.78 &      0.78 &     0.88 &      0.50 \\
LLAMA2-70B-chat       &       0.82 &    0.88 &      0.85 &     0.93 &      0.69 \\
Mistral-7B-instruct   &       0.80 &    0.69 &      0.74 &     0.84 &      0.41 \\
Mixtral-8x7B-instruct &       0.68 &    0.72 &      0.70 &     0.84 &      0.35 \\
\hline
\end{tabular}
\vspace{-2mm}

\end{table} 

In our study, we prioritized the multi-classification task for analyzing emergency situations, categorizing messages into specific emergency classes. We navigated the task's complexity due to the subtle and overlapping nature of class characteristics. For example, a user's message might pertain to 'fire,' 'electricity,' and 'shelter' simultaneously. We measured the models' performance using metrics such as accuracy, F1-score, and ROC-AUC, drawing from 1000 samples across all models. These metrics, detailed in Table \ref{tab:metrics}, are vital for a detailed evaluation of the models' capabilities in emergency context analysis.

In the domain of emergency classification, it is imperative to consider the importance of correctly identifying true emergencies (true positives) over incorrectly labeling non-emergencies as emergencies (false positives). Given the potentially life-threatening nature of missed or delayed emergency responses, the cost of false negatives is significantly higher than that of false positives. In such a context, Micro averaging is a crucial evaluation method, especially when dealing with class imbalances in multi-classification tasks. It combines the true positives, false positives, and false negatives across all classes by treating every prediction equally to compute metrics such as accuracy, F1 score, and ROC-AUC. This approach ensures that the metrics are not skewed by the overrepresentation of any single class, providing a more reliable measure of a model’s effectiveness in classifying emergency situations, and therefore, all metrics we presented are micro-averaged.

Accuracy, the simplest metric, represents the percentage of correct predictions. However, in a multi-class context with an unbalanced dataset, accuracy can be misleading. To address this, additional metrics were considered. The F1-score, used in binary classification, balances recall and precision. It is the harmonic mean of these two measures, providing a comprehensive assessment of the model's ability to identify each class. The Receiver Operating Characteristic (ROC) curve, crucial for binary classifiers, depicts performance across different discrimination thresholds. The Area Under the Curve (AUC) quantifies discriminative power, with higher values indicating superior performance. AUC values near 1 signify a perfect classifier, while 0.5 suggests random guessing. This metric is valuable for evaluating classifiers on imbalanced datasets, ensuring a thorough assessment in multi-classification tasks.

The LLAMA2 models, as illustrated in Table \ref{tab:metrics}, display a strong performance across accuracy, F1-score, and ROC-AUC, indicating their robust ability in emergency situation classification. The LLAMA2-70B model, in particular, excels, showcasing its strength in nuanced language processing. Conversely, the Mistral models, despite lower accuracy, show high F1 and ROC-AUC scores, suggesting effectiveness in identifying true positives over various thresholds. The LLAMA2 models also demonstrate superior capability in classifying non-English messages, providing accurate emergency analysis with efficient processing across multiple languages.

Based on our experimental outcomes, we have determined the LLAMA2 13B chat model to be well-suited for our objectives. This model stands out with its 13 billion parameters, striking an optimal balance between size and performance. It excels in deciphering the subtleties of language, which is paramount for analyzing social media communications during emergencies. Its architecture, fine-tuned for conversational AI, enables the delivery of coherent and context-aware responses, aligning perfectly with the critical need for precise and prompt interpretation of emergency-related data in our project.

\vspace{-2mm}
\section{Potential Problems}
The integration of artificial intelligence (AI) in emergency response services like 911 operations and healthcare has the potential to significantly enhance efficiency and predictive abilities. However, this integration requires careful consideration of various challenges. Misinformation in crisis situations is a primary concern, where AI can help identify and counter false information through data analysis and collaboration with fact-checking systems. Public awareness campaigns, supported by AI-generated content, are also vital for educating the public about misinformation.

Another significant issue is the potential for AI systems to exhibit societal biases, such as slower response times in certain demographic areas or misidentification in facial recognition technologies. This could lead to discriminatory practices. The reliability and accuracy of AI in high-stakes situations, like firefighting, are crucial, as AI systems based on imperfect data can make erroneous decisions. The vulnerability of these systems to cyberattacks adds to the risk.

The dilemma of balancing human judgment with AI decision-making is also critical, especially since AI lacks nuanced understanding of human emotions and ethical considerations. This could lead to AI systems prioritizing efficiency over compassionate responses. Additionally, issues of accessibility and equity arise, as the unequal distribution of technology can exacerbate social disparities.

The paramount importance in the application of AI within emergency settings lies in addressing privacy concerns, given the frequent processing of sensitive personal data. Ensuring data privacy while leveraging AI for public safety is a complex but essential task. Lastly, the transparency and explainability of AI systems are key to maintaining public trust and accountability, necessitating the development of AI models that are not only effective but also transparent in their decision-making processes.

\vspace{-2mm}
\section{Addressing Technical Challenges: Data and Model Poisoning in AI Systems}
In the realm of AI-driven emergency response systems, addressing the technical challenges posed by data and model poisoning is paramount for ensuring system integrity and reliability. 

\vspace{-2mm}
\subsection{Data Poisoning: Threats and Mitigation Strategies}
Data poisoning presents a formidable challenge to the integrity of AI models, notably LLMs. This phenomenon occurs when malicious entities intentionally introduce corrupted, misleading, or biased data into the training dataset. The objective is to manipulate the model's learning trajectory, resulting in outputs that are inaccurate, biased, or inappropriate. In emergency response scenarios, the ramifications of data poisoning can be profound, potentially leading to misdirection of services, erroneous interpretation of emergencies, or failure to recognize legitimate distress signals.

To counteract data poisoning, implementing stringent data validation and filtration protocols is essential \cite{cina2023wild}. This approach encompasses comprehensive scrutiny of data origins, continuous surveillance of incoming data for signs of manipulation, and the enforcement of rigorous data governance principles. Also, using adversarial training methods, which expose the model to made-up attack vectors while it is being trained, can make it much more resistant to these kinds of threats.

\vspace{-2mm}
\subsection{Model Poisoning: Identification and Countermeasures}
Model poisoning poses another significant threat but operates by targeting the AI model's learning mechanism or output directly. Attackers may employ tactics such as backdoor attacks, wherein specific inputs are engineered to trigger aberrant or detrimental model behaviors, or they might exploit vulnerabilities inherent in the model's architecture or its operational framework.

To safeguard against model poisoning, it is vital to conduct thorough audits of the model's architecture and deployment environment \cite{aljanabi2023safeguarding,mokander2023auditing}. This includes identifying potential backdoors and vulnerabilities that could be exploited by attackers. Regular updates and patches to the model's framework, along with continuous monitoring for unusual model behavior, are crucial steps in maintaining the security and integrity of AI systems in emergency response scenarios. Additionally, leveraging advanced defensive techniques, such as anomaly detection algorithms and reinforced learning environments, can provide an additional layer of protection against model poisoning attacks.

\vspace{-2mm}
\section{Conclusion \& Future Research}

In the realm of emergency response, LLMs, such as LLAMA2 and Mistral, can be transformative tools. Their integration into crisis management systems exemplifies the pivotal role of AI in enhancing efficiency and effectiveness. In this paper, we developed two advanced frameworks for \textit{optimizing 911 dispatch operations} and \textit{encouraging public cooperation}, both of which underscore the remarkable capabilities of LLMs. To this end, we have fine-tuned 5 different models: the LLAMA2 models 7B, 13B, 70B and the Mistral models 7B and 8x7B. After careful consideration of the available options, we ultimately chose LLAMA2 13B as our preferred solution. These models excel in their ability to process complex, multilingual communications, providing context-aware responses. This feature not only fosters inclusivity but also bridges language gaps, ensuring that diverse voices are heard and valued in challenging situations. However, it is crucial to acknowledge and address potential issues such as bias and privacy concerns in AI applications. The ethical and responsible deployment of AI is imperative in sensitive domains like crisis management. 

In summary, generative AI stands at the forefront of revolutionizing emergency response strategies. The ongoing refinement and evaluation of these models are crucial to enhance their effectiveness. Future research endeavors should concentrate on the ethical integration of AI, with a focus on fostering collaborative, equitable, and efficient crisis management practices. This approach is essential for safeguarding public well-being in times of crisis.

\bibliographystyle{IEEEtran}
\vspace{-2.5mm}
\bibliography{paper}

\begin{thebibliography}{10}
\providecommand{\url}[1]{#1}
\csname url@samestyle\endcsname
\providecommand{\newblock}{\relax}
\providecommand{\bibinfo}[2]{#2}
\providecommand{\BIBentrySTDinterwordspacing}{\spaceskip=0pt\relax}
\providecommand{\BIBentryALTinterwordstretchfactor}{4}
\providecommand{\BIBentryALTinterwordspacing}{\spaceskip=\fontdimen2\font plus
\BIBentryALTinterwordstretchfactor\fontdimen3\font minus \fontdimen4\font\relax}
\providecommand{\BIBforeignlanguage}[2]{{%
\expandafter\ifx\csname l@#1\endcsname\relax
\typeout{** WARNING: IEEEtran.bst: No hyphenation pattern has been}%
\typeout{** loaded for the language `#1'. Using the pattern for}%
\typeout{** the default language instead.}%
\else
\language=\csname l@#1\endcsname
\fi
#2}}
\providecommand{\BIBdecl}{\relax}
\BIBdecl

\bibitem{urban_systems_vulnerability}
A.~Team, ``Climate change and infrastructure, urban systems, and vulnerabilities: Technical report for the u.s. department of energy in support of the national climate assessment,'' \emph{National Climate Assessment}, 2014.

\bibitem{nyc_hurricane_sandy_2023}
{New York City Government}, ``Hurricane sandy,'' \url{https://www.nyc.gov/site/cdbgdr/hurricane-sandy/hurricane-sandy.page}, 2023.

\bibitem{pourebrahim2019understanding}
N.~Pourebrahim, S.~Sultana, J.~Edwards, A.~Gochanour, and S.~Mohanty, ``Understanding communication dynamics on twitter during natural disasters: A case study of hurricane sandy,'' \emph{International journal of disaster risk reduction}, vol.~37, p. 101176, 2019.

\bibitem{wang2019vulnerable}
Z.~Wang, N.~S. Lam, N.~Obradovich, and X.~Ye, ``Are vulnerable communities digitally left behind in social responses to natural disasters? an evidence from hurricane sandy with twitter data,'' \emph{Applied geography}, vol. 108, pp. 1--8, 2019.

\bibitem{hughes2014online}
A.~L. Hughes, L.~A. St.~Denis, L.~Palen, and K.~M. Anderson, ``Online public communications by police \& fire services during the 2012 hurricane sandy,'' in \emph{Proceedings of the SIGCHI conference on human factors in computing systems}, 2014, pp. 1505--1514.

\bibitem{loreti2022local}
S.~Loreti, E.~Ser-Giacomi, A.~Zischg, M.~Keiler, and M.~Barthelemy, ``Local impacts on road networks and access to critical locations during extreme floods,'' \emph{Scientific reports}, vol.~12, no.~1, p. 1552, 2022.

\bibitem{grace2021overcoming}
R.~Grace, ``Overcoming barriers to social media use through multisensor integration in emergency management systems,'' \emph{International Journal of Disaster Risk Reduction}, vol.~66, p. 102636, 2021.

\bibitem{misra2020information}
S.~Misra, P.~Roberts, and M.~Rhodes, ``Information overload, stress, and emergency managerial thinking,'' \emph{International Journal of Disaster Risk Reduction}, vol.~51, p. 101762, 2020.

\bibitem{andreassen2020information}
N.~Andreassen, O.~J. Borch, and A.~K. Sydnes, ``Information sharing and emergency response coordination,'' \emph{Safety Science}, vol. 130, p. 104895, 2020.

\bibitem{el2019case}
Z.~El~Khaled and H.~Mcheick, ``Case studies of communications systems during harsh environments: A review of approaches, weaknesses, and limitations to improve quality of service,'' \emph{International journal of distributed sensor networks}, vol.~15, no.~2, 2019.

\bibitem{clayman2023dispatching}
S.~E. Clayman and H.~Kevoe-Feldman, ``Dispatching first responders: Language practices and the dispatcher’s operational role in radio encounters with police officers,'' \emph{Discourse \& Society}, 2023.

\bibitem{dimou2021faster}
A.~Dimou, D.~G. Kogias, P.~Trakadas, F.~Perossini, M.~Weller, O.~Balet, C.~Z. Patrikakis, T.~Zahariadis, and P.~Daras, ``Faster: First responder advanced technologies for safe and efficient emergency response,'' in \emph{Technology Development for Security Practitioners}.\hskip 1em plus 0.5em minus 0.4em\relax Springer, 2021, pp. 447--460.

\bibitem{mondal2021emergency}
T.~Mondal, S.~Pramanik, P.~Pramanik, K.~N. Datta, P.~S. Paul, S.~Saha, and S.~Nandi, ``Emergency communication and use of ict in disaster management,'' \emph{Emerging technologies for disaster resilience: Practical cases and theories}, pp. 161--197, 2021.

\bibitem{adam2022mitigating}
H.~Adam, A.~Balagopalan, E.~Alsentzer, F.~Christia, and M.~Ghassemi, ``Mitigating the impact of biased artificial intelligence in emergency decision-making,'' \emph{Communications Medicine}, vol.~2, no.~1, p. 149, 2022.

\bibitem{10.1145/3375627.3375833}
H.~Lakkaraju and O.~Bastani, ``"how do i fool you?": Manipulating user trust via misleading black box explanations,'' in \emph{Proceedings of the AAAI/ACM Conference on AI, Ethics, and Society}.\hskip 1em plus 0.5em minus 0.4em\relax Association for Computing Machinery, 2020, p. 79–85.

\bibitem{poursabzi2021manipulating}
F.~Poursabzi-Sangdeh, D.~G. Goldstein, J.~M. Hofman, J.~W. Wortman~Vaughan, and H.~Wallach, ``Manipulating and measuring model interpretability,'' in \emph{Proceedings of the 2021 CHI conference on human factors in computing systems}, 2021, pp. 1--52.

\bibitem{weiss2021business}
J.~W. Weiss, \emph{Business ethics: A stakeholder and issues management approach}.\hskip 1em plus 0.5em minus 0.4em\relax Berrett-Koehler Publishers, 2021.

\bibitem{national2007improving}
N.~R. Council \emph{et~al.}, \emph{Improving disaster management: the role of IT in mitigation, preparedness, response, and recovery}.\hskip 1em plus 0.5em minus 0.4em\relax National Academies Press, 2007.

\bibitem{buchanan2022role}
P.~R. Buchanan and C.~Sparagowski, ``The role of emerging technologies and social justice in emergency management practice: The good, the bad, and the future,'' \emph{Justice, Equity, and Emergency Management}, pp. 175--199, 2022.

\bibitem{pilemalm2022barriers}
S.~Pilemalm, ``Barriers to digitalized co-production: the case of volunteer first responders,'' in \emph{19th International Conference on Information Systems for Crisis Response and Management, Tarbes, France, May 22-25, 2022}, 2022.

\bibitem{fcc_emergency_communications_2023}
F.~C. Commission, ``Emergency communications,'' \url{https://www.fcc.gov/general/emergency-communications}, 2023.

\bibitem{fcc_wireless_emergency_alerts_2023}
{Federal Communications Commission}, ``Wireless emergency alerts (wea),'' \url{https://www.fcc.gov/emergency-alert-system}, 2023.

\bibitem{sutton2021longitudinal}
J.~Sutton, Y.~Rivera, T.~K. Sell, M.~B. Moran, D.~Bennett~Gayle, M.~Schoch-Spana, E.~K. Stern, and D.~Turetsky, ``Longitudinal risk communication: A research agenda for communicating in a pandemic,'' \emph{Health Security}, vol.~19, no.~4, pp. 370--378, 2021.

\bibitem{stern2022critical}
E.~Stern and B.~Nussbaum, ``Critical infrastructure disruption and crisis management,'' in \emph{Oxford Research Encyclopedia of Politics}, 2022.

\bibitem{harrison2018improving}
K.~Harrison, ``Improving information sharing in the nyc emergency response community,'' \emph{Homeland Security Affairs}, 2018.

\bibitem{rane2023contribution}
N.~L. Rane, A.~Tawde, S.~P. Choudhary, and J.~Rane, ``Contribution and performance of chatgpt and other large language models (llm) for scientific and research advancements: a double-edged sword,'' \emph{International Research Journal of Modernization in Engineering Technology and Science}, vol.~5, no.~10, pp. 875--899, 2023.

\bibitem{cdc_global_health_2023}
{Centers for Disease Control and Prevention}, ``Working together to achieve improved emergency response around the world,'' 2023.

\bibitem{dhs_fire_risk_2023}
{Department of Homeland Security}, ``Dhs prepares and equips communities to address increased risk of fires,'' 2023.

\bibitem{innovative_emergency_management_2023}
{Online Masters in Public Health}, ``Innovative emergency management and response,'' \emph{USC Online MPH Program}, November 2023.

\bibitem{chen2023early}
W.~Chen, G.~Rao, D.~Kang, Z.~Wan, and D.~Wang, ``Early report of the source characteristics, ground motions, and casualty estimates of the 2023 m w 7.8 and 7.5 turkey earthquakes,'' \emph{Journal of Earth Science}, 2023.

\bibitem{touvron2023llama}
H.~Touvron, L.~Martin, K.~Stone, P.~Albert, A.~Almahairi, Y.~Babaei, N.~Bashlykov, S.~Batra, P.~Bhargava, S.~Bhosale \emph{et~al.}, ``Llama 2: Open foundation and fine-tuned chat models,'' \emph{arXiv preprint arXiv:2307.09288}, 2023.

\bibitem{mistralai_mixtral_8x7b}
{Mistral AI}, ``Mixtral 8x7b,'' \url{https://mistral.ai/}.

\bibitem{us_census_bureau_language_use_2019}
{U.S. Census Bureau}, ``Language use in the united states: 2019,'' American Community Survey, 2019.

\bibitem{Welch22}
J.~P. Welch, ``The challenges of public service organizations in emergency, crisis, and disaster management,'' in \emph{Crisis Management}, 2022.

\bibitem{jain2023neftune}
N.~Jain, P.~yeh Chiang, Y.~Wen, J.~Kirchenbauer, H.-M. Chu, G.~Somepalli, B.~R. Bartoldson, B.~Kailkhura, A.~Schwarzschild, A.~Saha, M.~Goldblum, J.~Geiping, and T.~Goldstein, ``Neftune: Noisy embeddings improve instruction finetuning,'' 2023.

\bibitem{dao2023flashattention2}
T.~Dao, ``Flashattention-2: Faster attention with better parallelism and work partitioning,'' 2023.

\bibitem{cina2023wild}
A.~E. Cin{\`a}, K.~Grosse, A.~Demontis, S.~Vascon, W.~Zellinger, B.~A. Moser, A.~Oprea, B.~Biggio, M.~Pelillo, and F.~Roli, ``Wild patterns reloaded: A survey of machine learning security against training data poisoning,'' \emph{ACM Computing Surveys}, 2023.

\bibitem{aljanabi2023safeguarding}
M.~Aljanabi, ``Safeguarding connected health: Leveraging trustworthy ai techniques to harden intrusion detection systems against data poisoning threats in iomt environments,'' \emph{Babylonian Journal of Internet of Things}, 2023.

\bibitem{mokander2023auditing}
J.~M{\"o}kander, J.~Schuett, H.~R. Kirk, and L.~Floridi, ``Auditing large language models: a three-layered approach,'' \emph{AI and Ethics}, pp. 1--31, 2023.

\end{thebibliography}

\end{document}